\documentclass[10pt,twocolumn,letterpaper]{article}

\usepackage{cvpr}              
\usepackage{multirow}

\providecommand{\degree}{$^\circ$}

\definecolor{cvprblue}{rgb}{0.21,0.49,0.74}
\usepackage[pagebackref,breaklinks,colorlinks,allcolors=cvprblue]{hyperref}

\def\paperID{48} 
\def\confName{CVPR}
\def\confYear{2026}

\title{One Perturbation, Two Failure Modes: Probing VLM Safety via Embedding-Guided Typographic Perturbations}

\author{Ravikumar Balakrishnan\\
Cisco Systems\\
{\tt\small ravikuba@cisco.com}
\and
Sanket Mendapara\\
Cisco Systems\\
{\tt\small sanket@cisco.com}
}

\begin{document}
\maketitle
\begin{abstract}
Typographic prompt injection exploits vision language models' (VLMs) ability to read text rendered in images, posing a growing threat as VLMs power autonomous agents. Prior work typically focus on maximizing attack success rate (ASR) but does not explain \emph{why} certain renderings bypass safety alignment. We make two contributions. First, an empirical study across four VLMs including GPT-4o and Claude, twelve font sizes, and ten transformations reveals that multimodal embedding distance strongly predicts ASR ($r{=}{-}0.71$ to ${-}0.93$, $p{<}0.01$), providing an interpretable, model agnostic proxy. Since embedding distance predicts ASR, reducing it should improve attack success, but the relationship is mediated by two factors: perceptual readability (whether the VLM can parse the text) and safety alignment (whether it refuses to comply). Second, we use this as a red teaming tool: we directly maximize image text embedding similarity under bounded $\ell_\infty$ perturbations via CWA-SSA across four surrogate embedding models, stress testing both factors without access to the target model. Experiments across five degradation settings on GPT-4o, Claude Sonnet 4.5, Mistral-Large-3, and Qwen3-VL confirm that optimization recovers readability and reduces safety aligned refusals as two co-occurring effects, with the dominant mechanism depending on the model's safety filter strength and the degree of visual degradation.
\end{abstract}
\section{Introduction}
\label{sec:intro}

\begin{figure*}[t]
\centering
\includegraphics[width=0.95\textwidth]{figures/method_figure.png}
\caption{Overview of our embedding-guided adversarial optimization. A degraded typographic image (left) is optimized via the popular CWA-SSA (Common Weakness Attack with Spectral Simulation Augmentation) across four surrogate embedding models to maximize cosine similarity with the source prompt text embedding (top). The resulting image (right) is visually similar but semantically realigned. This single objective produces two co-occurring effects: readability recovery and reduction of safety-aligned refusals, whose relative contribution varies across models and degradation settings.}
\label{fig:method}
\end{figure*}

Vision-language models (VLMs) can read and follow textual instructions embedded within images, creating an attack vector known as \emph{typographic prompt injection} (TPI)~\cite{gong2023figstep}. By rendering harmful prompts as images, attackers bypass text-based safety filters because the malicious instruction arrives through the visual modality. In practice, the attack surface is heterogeneous: adversarial text may appear at varying font sizes, under visual transformations (rotation, blur, noise), and across different attack strategies, while VLMs vary substantially in vulnerability. Yet prior work focuses on maximizing attack success rate (ASR) through novel attack designs~\cite{gong2023figstep,wang2025typographic,cao2025scenetap}, without characterizing \emph{why} certain renderings bypass safety alignment while others do not. Building this understanding is a prerequisite for designing robust VLM safety mechanisms.

This paper makes two contributions. First, we present a systematic empirical study (\cref{sec:empirical}) showing that multimodal embedding distance is an \emph{interpretable, model-agnostic indicator} of typographic attack success ($r = -0.71$ to $-0.93$). Second, we use this insight as a red teaming tool (\cref{sec:method}): we directly maximize image-text embedding similarity under bounded perturbations, probing the robustness of VLM safety alignment. Our experiments (\cref{sec:experiments}) reveal that this single objective exposes two co-occurring vulnerabilities: recovering readability and reducing safety aligned refusals. The relative contribution depends on the model's safety filter strength and the degree of visual degradation.

\section{Text-Image Embedding Distance Correlates with ASR}
\label{sec:empirical}

Before designing an optimization method, we first need to understand what makes a typographic attack succeed or fail. This section presents a controlled empirical study characterizing how rendering conditions (font size, visual transformations) affect attack success, and tests whether off-the-shelf multimodal embedding models can predict this variation. We evaluate 1{,}000 prompts curated from SALAD-Bench~\cite{li2024salad} across four VLMs, namely, GPT-4o, Claude Sonnet 4.5, Mistral-Large-3, and Qwen3-VL-4B. Each prompt is rendered as black text on a white background at $1024 \times 1024$ resolution using a standard sans serif font with word wrapping, and tested both as raw text and as typographic images at twelve font sizes (6--28px), as well as under ten visual transformations applied to 20px renderings. Attack success is judged by GPT-4o using a standardized evaluation protocol.

\paragraph{Font size and transformation effects.}
Font size significantly affects ASR (\cref{tab:conditions}). Very small fonts (6px) reduce ASR across all models, though the magnitude varies: GPT-4o drops to 0.3\% and Claude to 1.2\%, while Mistral retains 15.0\% and Qwen3-VL retains 23.9\%. ASR increases rapidly from 6px to 10px and then more gradually beyond that. Among visual transformations applied to 20px renderings, heavy blur and triple degradation (blur + noise + low contrast) reduce ASR most severely, while rotation affects models asymmetrically: Mistral and Qwen3-VL drop substantially under 90\degree{} rotation, whereas GPT-4o and Claude are relatively stable.

\begin{table}[t]
\caption{ASR (\%) and JinaCLIP L2 distance across rendering conditions. Font sizes are rendered directly; transformations are applied to 20px text. $n{=}1{,}000$.}
\label{tab:conditions}
\vspace{-2mm}
\centering\small
\begin{tabular}{lccccc}
\toprule
Condition & $d$ & GPT-4o & Claude & Mistral & Qwen \\
\midrule
6px  & 1.265 & 0.3  & 1.2  & 15.0 & 23.9 \\
8px  & 1.242 & 3.5  & 10.5 & 67.9 & 40.8 \\
10px & 1.192 & 6.4  & 21.6 & 73.5 & 43.1 \\
20px & 1.115 & 7.7  & 16.4 & 73.8 & 48.2 \\
\textit{Text} & -- & \textit{35.6} & \textit{46.6} & \textit{85.0} & \textit{48.9} \\
\midrule
Rot 90\degree & 1.173 & 4.8 & 9.2 & 39.9 & 18.2 \\
Triple deg. & 1.227 & 3.0 & 3.2 & 44.6 & 28.7 \\
Heavy blur & 1.244 & 2.7 & 0.7 & 26.7 & 25.2 \\
\bottomrule
\end{tabular}
\vspace{-3mm}
\end{table}

\paragraph{Embedding distance correlates with ASR.}
For each prompt $p$ rendered as image $I$, we compute the L2 distance between normalized multimodal embeddings:
\begin{equation}
d(p, I) = \left\| \frac{\text{ImgEnc}(I)}{\|\text{ImgEnc}(I)\|} - \frac{\text{TxtEnc}(p)}{\|\text{TxtEnc}(p)\|} \right\|_2
\label{eq:l2dist}
\end{equation}
using JinaCLIP~\cite{koukounas2024jina} (1024-d) and Qwen3-VL-Embedding (2048-d). \cref{tab:conditions} already hints at this relationship: conditions that increase L2 distance tend to reduce ASR. We confirm this quantitatively in \cref{tab:corr}. Both embedding models show strong negative correlations with ASR across all four VLMs ($r = -0.71$ to $-0.93$ for font sizes, $r = -0.72$ to $-0.99$ for transformations, nearly all $p < 0.01$).

\begin{table}[t]
\caption{Pearson correlation between L2 embedding distance and ASR across font sizes (left) and visual transformations (right).}
\label{tab:corr}
\vspace{-2mm}
\centering\small
\begin{tabular}{lcccc}
\toprule
 & \multicolumn{2}{c}{Font sizes} & \multicolumn{2}{c}{Transforms} \\
\cmidrule(lr){2-3} \cmidrule(lr){4-5}
Emb.\ / VLM & $r$ & $p$ & $r$ & $p$ \\
\midrule
Jina / GPT-4o & $-0.795$ & $.002$ & $-0.829$ & $.003$ \\
Jina / Claude & $-0.725$ & $.008$ & $-0.893$ & $.001$ \\
Jina / Mistral & $-0.714$ & $.009$ & $-0.805$ & $.005$ \\
Jina / Qwen & $-0.799$ & $.002$ & $-0.717$ & $.020$ \\
\midrule
Qwen / GPT-4o & $-0.843$ & $.001$ & $-0.628$ & $.052$ \\
Qwen / Claude & $-0.812$ & $.001$ & $-0.880$ & $.001$ \\
Qwen / Mistral & $-0.899$ & ${<}.001$ & $-0.987$ & ${<}.001$ \\
Qwen / Qwen & $-0.934$ & ${<}.001$ & $-0.965$ & ${<}.001$ \\
\bottomrule
\end{tabular}
\vspace{-3mm}
\end{table}

\paragraph{Key insight.}
The strong correlation between embedding distance and ASR implies that reducing the distance between a typographic image and its source text in embedding space should increase attack success. Visually degraded renderings reside in high distance regions and present the most room for improvement. However, the relationship between embedding distance and ASR is mediated by two factors: perceptual readability (whether the VLM can parse the text) and safety alignment (whether the VLM refuses to comply). Both are potential targets for an optimization that reshapes the image's representation in embedding space. This motivates our approach.

\section{Method: Probing Safety Robustness via Embedding Guided Optimization}
\label{sec:method}

\subsection{Core Idea}

The empirical results in \cref{sec:empirical} suggest that image-text embedding distance strongly correlates with typographic attack success: when a typographic image is close to its source text in embedding space, VLMs are more likely to comply with the embedded instruction. A natural question is whether this relationship is causal: does \emph{reducing} the embedding distance of a degraded typographic image increase its attack success, and if so, what does this reveal about VLM safety mechanisms?

To answer this, we construct a red teaming procedure that applies a bounded perturbation to a degraded typographic image, reducing its embedding distance to the source text and pushing it from a high distance (low ASR) region into a low distance (high ASR) region. This allows us to stress test VLM safety alignment under controlled conditions where the visual appearance of the image remains largely unchanged.

We hypothesize that this perturbation can improve ASR through two mechanisms, each exposing a different fragility. First, when the VLM cannot read the degraded text, the perturbation may recover semantic content in the model's internal representation without restoring visual legibility, revealing that safety which relies on the text being unreadable is brittle. Second, when the VLM can already read the text but refuses to comply, the perturbation alters the holistic image representation that feeds into the model's safety reasoning, exposing cases where alignment is not robust to small input perturbations.

Critically, this optimization does not require access to the target VLM, its safety classifier, or any class labels. The text embedding of the attack prompt serves as a fixed target, and the objective is to make the image ``look more like the text'' to a set of surrogate embedding models.

\subsection{Optimization Objective}
\label{sec:objective}

Given a set of $K$ multimodal embedding models $\{f_k\}_{k=1}^K$, an initial typographic image $I_0$ encoding prompt $p$, and the corresponding text embeddings $\mathbf{t}_k = \text{TxtEnc}_k(p)$, we seek:
\begin{equation}
\max_{\delta:\,\|\delta\|_\infty \leq \epsilon} \; \frac{1}{K} \sum_{k=1}^{K} \cos\!\left(\text{ImgEnc}_k(I_0 + \delta),\; \mathbf{t}_k\right)
\label{eq:objective}
\end{equation}
where $\cos(\cdot,\cdot)$ denotes cosine similarity and $\epsilon$ bounds the $\ell_\infty$ perturbation. The text embeddings $\mathbf{t}_k$ are computed once per prompt and remain fixed throughout optimization.

\subsection{Optimization Procedure}

We optimize over an ensemble of $K{=}4$ multimodal embedding models spanning contrastive and generative families: Qwen3-VL-Embedding-2B, JinaCLIP v2~\cite{koukounas2024jina}, OpenAI CLIP ViT-L/14-336, and SigLIP SO400M. This architectural diversity encourages perturbations that capture general features of text readability rather than artifacts of a single encoder.

To solve \cref{eq:objective}, we adapt the Common Weakness Attack with Spectral Simulation Augmentation (CWA-SSA)~\cite{chen2024rethinking}, originally designed for transferable adversarial attacks on classifiers. CWA-SSA's sequential inner loop processes each ensemble model in turn, finding perturbation directions that benefit \emph{all} models, while its spectral and input diversity augmentations encourage convergence to flat loss basins that transfer to held out models. Prior work applied CWA to maximize image embedding similarity to a reference image, disrupting image understanding~\cite{chen2024rethinking,dong2023bard}. We instead maximize similarity between a degraded typographic image and its source \emph{prompt text}, pushing it into a high compliance region of embedding space.

\section{Experiments}
\label{sec:experiments}

\subsection{Setup}

\paragraph{Degradation settings.}
We evaluate the robustness of VLM safety alignment under five rendering conditions that produce low baseline ASR:
\begin{itemize}[nosep,leftmargin=*]
    \item \textbf{6px font}: very small text rendered at 6px on 1024$\times$1024 images (baseline ASR: 0.3--24\% across models).
    \item \textbf{8px font}: small text at 8px (baseline ASR: 3.5--67.9\%).
    \item \textbf{90\degree{} rotation}: 20px text rotated 90\degree{} (baseline ASR: 4.8--39.9\%).
    \item \textbf{Triple degradation}: blur + noise + low contrast applied to 20px text (baseline ASR: 3.0--44.6\%).
    \item \textbf{Heavy blur}: $\sigma{=}5$ Gaussian blur on 20px text (baseline ASR: 0.7--26.7\%).
\end{itemize}

\paragraph{Optimization hyperparameters.}
All experiments use: 4 model ensemble (Qwen3-VL-Embedding-2B, JinaCLIP v2, OpenAI CLIP ViT-L/14-336, SigLIP SO400M), $T{=}100$ steps, $\epsilon{=}32/255$ ($\ell_\infty$), inner step $\alpha_{\text{in}}{=}5/255$, outer step $\alpha_{\text{out}}{=}4/255$, momentum $\mu{=}0.9$, SSA samples $S{=}3$ with $\sigma{=}16$, $\rho{=}0.5$, and DI probability 0.7.

\paragraph{Sample selection.}
We select 50 samples where the text attack succeeds on both GPT-4o and Claude, the two models with the lowest baseline image ASR, but the degraded image fails on both. Among these, we rank by how readily the 20px baseline recovers attack success and take the top 50.

\paragraph{Evaluation.}
We evaluate ASR on GPT-4o, Claude Sonnet 4.5, Mistral-Large-3, and Qwen3-VL-4B using a GPT-4o based judge consistent with the empirical study.

\subsection{Results}

\begin{table}[t]
\caption{ASR before (B) and after (A) embedding guided optimization on 50 selected samples. Bold = largest gain per setting.}
\label{tab:results_all}
\vspace{-2mm}
\centering\footnotesize
\setlength{\tabcolsep}{3pt}
\begin{tabular}{l rrrr r}
\toprule
Setting & GPT-4o & Claude & Mistral & Qwen & Best $\Delta$ \\
\midrule
Rot 90\degree\ B  & 0\%  & 0\%  & 78\% & 40\% & \\
Rot 90\degree\ A  & 16\% & \textbf{22\%} & 72\% & 54\% & \textbf{+22\%} \\
\midrule
6px B             & 0\%  & 0\%  & 44\% & 54\% & \\
6px A             & 2\%  & 2\%  & \textbf{58\%} & 64\% & \textbf{+14\%} \\
\midrule
8px B             & 0\%  & 0\%  & 88\% & 68\% & \\
8px A             & \textbf{10\%} & 8\%  & 82\% & 64\% & \textbf{+10\%} \\
\midrule
Triple Deg.\ B    & 0\%  & 0\%  & 86\% & 76\% & \\
Triple Deg.\ A    & 8\%  & \textbf{14\%} & 74\% & 64\% & \textbf{+14\%} \\
\midrule
Heavy Blur B      & 0\%  & 0\%  & 52\% & 44\% & \\
Heavy Blur A      & 10\% & \textbf{28\%} & 72\% & 54\% & \textbf{+28\%} \\
\bottomrule
\end{tabular}
\vspace{-3mm}
\end{table}

\paragraph{Failure mode analysis.}
To diagnose where gains originate, we classify each failure as an explicit refusal, a readability failure (model reports it cannot read the text), or other (misreading or tangential response). Three patterns emerge across settings.

\textbf{Readability recovery dominates at 6px and heavy blur.} At 6px, 35 of GPT-4o's 50 pre-optimization failures are readability related, with only 10 refusals. After optimization, readability failures drop from 35 to 10, but refusals rise from 10 to 28, yielding only 1 success: the perturbation makes the text readable, but GPT-4o's safety filter catches the now legible requests. Claude on heavy blur shows the strongest overall gain (+28\%): empty responses, where Claude previously could not process the image at all, drop from 39 to 14 as the optimization makes the blurred content parseable. For Mistral, whose safety filter is weaker, readability gains translate directly to ASR: misreadings drop from 20 to 5 on heavy blur (+20\%) and from 23 to 11 on 6px (+14\%).

\textbf{Mixed readability and refusal reduction at 90\degree rotated (rot90) and 8px.} At rot90, GPT-4o has 28 refusals but also 12 readability failures, indicating that 20px rotated text is not fully legible. After optimization, readability failures drop (12$\to$7) and 8 successes emerge (+16\%). Claude gains +22\%, with refusals dropping from 27 to 19 and empty responses from 13 to 8. At 8px, a similar pattern holds: GPT-4o and Claude gain +10\% and +8\% from a 0\% baseline.

\textbf{High baseline models can be hurt.} Models with already high baseline ASR sometimes show decreased performance after optimization. Mistral drops from 86\%$\to$74\% on triple degradation, with refusals increasing from 1 to 7, suggesting the perturbation can trigger safety mechanisms that were not previously engaged. This pattern repeats on rot90 ($-$6\%) and 8px ($-$6\%).

\section{Related work.}
Typographic prompt injection was introduced by FigStep~\cite{gong2023figstep} and extended to multi image~\cite{wang2025typographic}, scene coherent~\cite{cao2025scenetap}, and benchmark scale~\cite{ma2025scam} settings. CLIP based embeddings have been used for adversarial attacks~\cite{zhou2023advclip}, attack placement~\cite{wang2025typographic}, and defenses~\cite{azuma2023defense}. Our optimization builds on CWA-SSA~\cite{chen2024rethinking}, which achieves transferability via sequential ensemble updates and spectral augmentation; we adapt it to maximize image-text embedding similarity. We differ from prior work by using embedding alignment as both an \emph{explanatory tool} (predicting ASR) and a \emph{red teaming objective} (probing safety robustness).

\section{Conclusion}
\label{sec:conclusion}

We have shown that multimodal embedding distance is an interpretable predictor of typographic attack success ($r$ up to $-0.93$), and that optimizing this distance under a bounded perturbation exposes two co-occurring fragilities in VLM safety: readability recovery (Claude heavy blur: +28\%) and refusal reduction (Claude 90\degree rotated: +22\%). Which mechanism dominates depends on the degradation severity and the model's safety filter strength. The finding that bounded perturbations can reduce refusal rates without improving visual legibility points to a need for safety mechanisms robust in representation space, not only in the pixel domain.

\paragraph{Limitations.}
Our evaluation uses a single dataset (SALAD-Bench) and one rendering style (black text on white). The optimization requires surrogate embedding models and is computationally expensive (${\sim}2.5$h per sample). We do not evaluate against detection based defenses.

{
    \small
    \bibliographystyle{ieeenat_fullname}
    \bibliography{main}
}

\end{document}